\documentclass[journal,12pt,onecolumn,draftclsnofoot,]{IEEEtran}
\usepackage{pbox}
\usepackage{tikz}
\usepackage{pgfgantt}
\usepackage{subfigure}
\usepackage{caption}
\usepackage{adjustbox} 
\usepackage{amssymb}

\usepackage{xcolor}
\usepackage{afterpage}

\usepackage{soul}

\usepackage{amsmath}

\usepackage{subfigure}
\usepackage{tikz}  
\usetikzlibrary{fit}   
\usepackage{upgreek}
\usepackage{pgfgantt}

\usetikzlibrary{matrix,shapes,arrows,positioning,chains}
\tikzstyle{process} = [rectangle, minimum width=3cm, minimum height=1cm, text centered, draw=black]
\tikzstyle{arrow} = [thick,->,>=stealth]
\tikzstyle{io} = [trapezium, trapezium left angle=70, trapezium right angle=110, text centered]

\ifCLASSINFOpdf
\else
\fi
\usepackage{pbox}
\usepackage{tikz}
\usepackage{pgfgantt}
\usepackage{subfigure}
\usepackage{gensymb}
\usepackage{amsmath}

\usepackage{array}
\usepackage{graphicx}
\usepackage{multirow}
\usepackage{xcolor,colortbl}

\usetikzlibrary{matrix,shapes,arrows,positioning,chains}
\tikzstyle{process} = [rectangle, minimum width=3cm, minimum height=1cm, text centered, draw=black]
\tikzstyle{arrow} = [thick,->,>=stealth]
\tikzstyle{io} = [trapezium, trapezium left angle=70, trapezium right angle=110, text centered]

\makeatletter
\def\@copyrightspace{\relax}
\makeatother

\begin{document}
%
\title{Federated Transfer Learning: concept and applications}
%
%
%

\author{Sudipan~Saha
        and~Tahir~Ahmad
\thanks{Sudipan Saha is with Technical University of Munich, Taufkirchen, Germany and Tahir Ahmad is with Fondazione Bruno
Kessler, Trento, Italy. E-mail: sudipan.saha@tum.de, ahmad@fbk.eu}
}

\maketitle

\begin{abstract}
Development of Artificial Intelligence (AI) is inherently tied to the development of data. However, in most industries data exists in form of isolated islands, with limited scope of sharing between different organizations. This is an hindrance to the further development of AI. Federated learning has emerged as a possible solution to this problem in the last few years
without compromising user privacy. Among different variants of the federated learning, noteworthy is federated transfer learning (FTL) that allows knowledge to be transferred across domains that do not have many overlapping features and users. In this work we provide a comprehensive survey of the existing works on this topic. In more details, we study the background of FTL and its different existing applications. We further analyze FTL from privacy and machine learning perspective.
\end{abstract}

\begin{IEEEkeywords}
Federated Learning; Transfer Learning; Machine Learning; Privacy-preserving.
\end{IEEEkeywords}

\fboxsep=0mm
\fboxrule=0.1pt

%
\IEEEpeerreviewmaketitle

\bstctlcite{IEEEexample:BSTcontrol}

\section{Introduction}
\label{sec:intro}
Currently machine learning is playing an important role in many applications, including commodity recommendation, risk analysis, and image analysis. This is due to its excellent capability to extract insights from data. Machine learning and deep learning techniques strongly depend on the availability of data \cite{goodfellow2016deep}. Abundance of data has led to the overwhelming success of machine learning in image analysis \cite{deng2009imagenet}, remote
sensing \cite{saha2019unsupervisedIgarss, saha2020building}, and many other fields.
Traditionally, such machine learning or deep learning models are trained over a centralized corpus of data. 
 While it is easier to collect image data, they are also more intuitive to label without strong domain knowledge.
 Moreover, image data is generally not sensitive and shared without restriction among different parties. However, training data is not easy to obtain in some industries, e.g., finance
and healthcare \cite{jing2019quantifying}. Labeling data from such industries require strong professional expertise. Moreover, data in such industries are generally protected by different privacy and security related restrictions \cite{voigt2017eu}. Additionally, there exists practical risks of data abuse once the data are
shared to the third parties. In such industries, data exists in the form of isolated island \cite{yang2019federated}. Competition between different organizations also hinder the sharing and thus exposing one's user data to another. Thus,
such industries own insufficient data to train reliable machine learning frameworks.
\par
Federated Learning (FL) can be used to overcome the above-mentioned constraints by using data from different organizations to train machine learning model, however not violating the different data related regulations. FL system was first proposed by Google in 2016 \cite{mcmahan2016federated, konevcny2016federated, konevcny2016federatedOptimization}. Their method was proposed for 
mobile devices that enables users to train a centralized model while their data are stored locally.  
Thus, federated learning technique can be used prevent the leakage of private information. While centralized learning needs to collect data from users and store them in centralized server, federated learning can learn a  global model while the data are distributed on the users' devices.
Many other works adopted the federated learning framework \cite{zhu2019multi, lim2020federated}.
This led to emergence of sub-groups within federated learning, e.g., horizontal federated learning and vertical federated learning \cite{yang2019federated}.

\par
A constraint imposed by the traditional federated learning is that training data owned by different organizations need to share same feature space. In practice, this is never the case in industries like finance or healthcare. To mitigate this shortcoming, Federated Transfer Learning (FTL) was proposed \cite{liu2020secure}. 
Different participants in FTL can have their specific feature
space, thus making it suitable for practical scenarios. FTL takes inspiration from transfer learning, a paradigm already popular in image analysis \cite{saha2020unsupervised, saha2021ultrasound}. In this setting, machine learning models trained on a large dataset for one problem/domain is applied to a different but related problem/domain \cite{saha2019unsupervised}.  The performance of transfer learning is strongly dependent on interrelation between different domains. While talking about federated learning,  stakeholders in the same
data federation are usually organizations from the same industry. Thus, it is suitable to apply transfer learning in the federated learning framework.  
\par
Federated transfer learning lies at the intersection of two different but fast-evolving fields: machine learning and information privacy. Thus it is an imperative to bridge the gap between them to fully exploit the benefits of federated transfer learning. This motivated us to investigate into the different aspects related to federated transfer learning. It is important to understand how transfer learning and federated learning intermarried to give rise to federated transfer learning. It is critical to understand about how FTL has been applied in different real-life applications. It is important to understand the different privacy and machine learning aspects related to FTL. Keeping them in mind, in this paper, we present a comprehensive survey of the federated transfer learning. Towards this we: 1) outline the definition of FTL, 2) present some case studies on FTL, 3) analyze FTL from privacy aspect, and 4) analyze FTL from machine learning aspects.

\par
We briefly discuss about horizontal and vertical federated learning in Section \ref{sectionRelatedWork}. Federated transfer learning is defined in 
Section \ref{sectionFTLDefinition}. We detail the case studies on FTL in Section \ref{sectionFTLCaseStudies}.  We analyze FTL's privacy aspects in Section \ref{sectionFTLPrivacy}
and machine learning aspects in Section \ref{sectionFTLMachineLearning}. Datasets used in the FTL related works are briefly 
presented in Section \ref{sectionFTLDataset}. We conclude the work in Section \ref{sectionConclusion}.

\section{Related work}
\label{sectionRelatedWork}
For the scenario where datasets held by different users differ mostly in samples, federated learning can be categorized into horizontal and vertical federated learning. In this section we briefly review them. We also briefly review transfer learning, considering its relation to the
federated transfer learning.
\subsection{Horizontal federated learning}
\label{sectionRelatedWorkHorFedLearning}
Horizontal federated learning is a system in which all the parties share the same feature space.  However, their userbase may be significantly different. Such parties can collaboratively
learn a model with help of a server. Each party locally computes training gradient and masks them with
some encryption or privacy-preservation technique \cite{aono2017privacy}. All parties send encrypted gradient to the server. The server aggregates them and sends the aggregated result to all parties. Parties update their model with the decrypted aggregated gradient and this way all parties share final model parameters \cite{mcmahan2016federated}. An example of horizontal federated learning is a set of banks located in the same city or region and thus sharing same set of features, however very few common users. Horizontal federated learning can be used to learn a common model by agglomerating models learnt in individual banks.
The horizontal federal learning can be implemented with different machine learning algorithms without
changing the main framework. 

\subsection{Vertical federated learning}
\label{sectionRelatedWorkVerFedLearning}
In the vertical federated learning, participating parties do not expose users that do not overlap among the parties \cite{hardy2017private, yang2019federated}. Overlapping users are found by using an encryption-based user ID alignment.
Since different parties have different features corresponding to the common users, 
vertical federated learning aggregates different features from different parties and computes
the training loss and gradients in a privacy-preserving manner \cite{hardy2017private}. Subsequently computed gradients are
used to train the model. Vertical federated learning assumes honest participants and there is no hard requirement of a third party, as illustrated here \cite{yang2019parallel}. However, sometimes  to 
secure computations between the participants, an additional party is introduced \cite{yang2019federated}. An example of the vertical federated learning is case of cooperation between the online retailers and the insurers. They own their own feature space (and labels). However, their have significant amount of common users. In such cases, vertical federated learning exploits the situation by merging the features together to create a larger feature space for machine
learning tasks \cite{wang2019interpret}.

\subsection{Transfer learning}
\label{sectionRelatedWorkTransferLearning}
Most machine learning algorithms assume that the training data and the test data have same distribution and they are in the same feature space. However, this assumption does not hold in most real life scenario. In most practical cases, we have intend to analyze one domain of interest, while we have sufficient training data in another domain. As an example, we can consider the problem of classification of a 
product review \cite{pan2009survey}. Using traditional machine learning approach, sufficient number of product review needs to be collected and annotated for training.
The distribution of review data varies from product to product and hence this training data collection and annotation process needs to
be performed separately for each product. Separate data collection for each product is time-consuming and challenging. Transfer learning emerged
as a framework to address this problem. Transfer learning provides a mechanism of training model on one product and reusing it on another product. 
\par
Transfer learning allows the tasks, domains, and distributions in the training and testing to be different. Pan and Yang \cite{pan2009survey} noted that
research on transfer learning attracted more attention since 1995 and was tackled under different names, e.g., knowledge transfer, inductive
transfer, and multi-task learning. Different approaches were adopted for transfer learning \cite{pan2009survey}:
\begin{enumerate}
\item \textit{Instance transfer} re-weights labeled data in the source domain to reuse it in the target domain \cite{jiang2007instance}.
\item \textit{Parameter transfer} discovers shared parameters between the source and the target domain \cite{gao2008knowledge}.
\item \textit{Feature-representation transfer} finds a feature-representation such that it reduces the difference between source and target domains \cite{argyriou2008spectral}.
\item \textit{Relational-knowledge transfer} builds mapping of relational knowledge between source and target domain \cite{mihalkova2007mapping}.
\end{enumerate}
In the last decade, deep learning emerged as a very successful paradigm in the machine learning research. However, deep learning is even more
data dependent than the previous machine learning algorithms. To tackle this, researchers started adopting deep transfer learning, to utilize knowledge from other fields by
deep neural networks \cite{tan2018survey}. In addition to the four approaches defined above, another approach that gained significant attention in the deep transfer learning is 
adversarial-based deep transfer learning that introduces adversarial techniques based on generative adversarial network (GAN) \cite{goodfellow2014generative} and its variants to find transferable representations applicable to both the source domain and the target domain.
Deep transfer learning is closely related to other topics in the deep learning, e.g., deep domain adaptation \cite{saha2019unsupervisedIgarss}.
One important line of investigation in the deep transfer learning is that which networks are more suitable for transfer and which features are transferable in the deep network \cite{yosinski2014transferable}.

\section{Federated Transfer Learning}
\label{sectionFTLDefinition}

\begin{figure}[ht]
\centering
\includegraphics[width=0.8\linewidth,keepaspectratio]{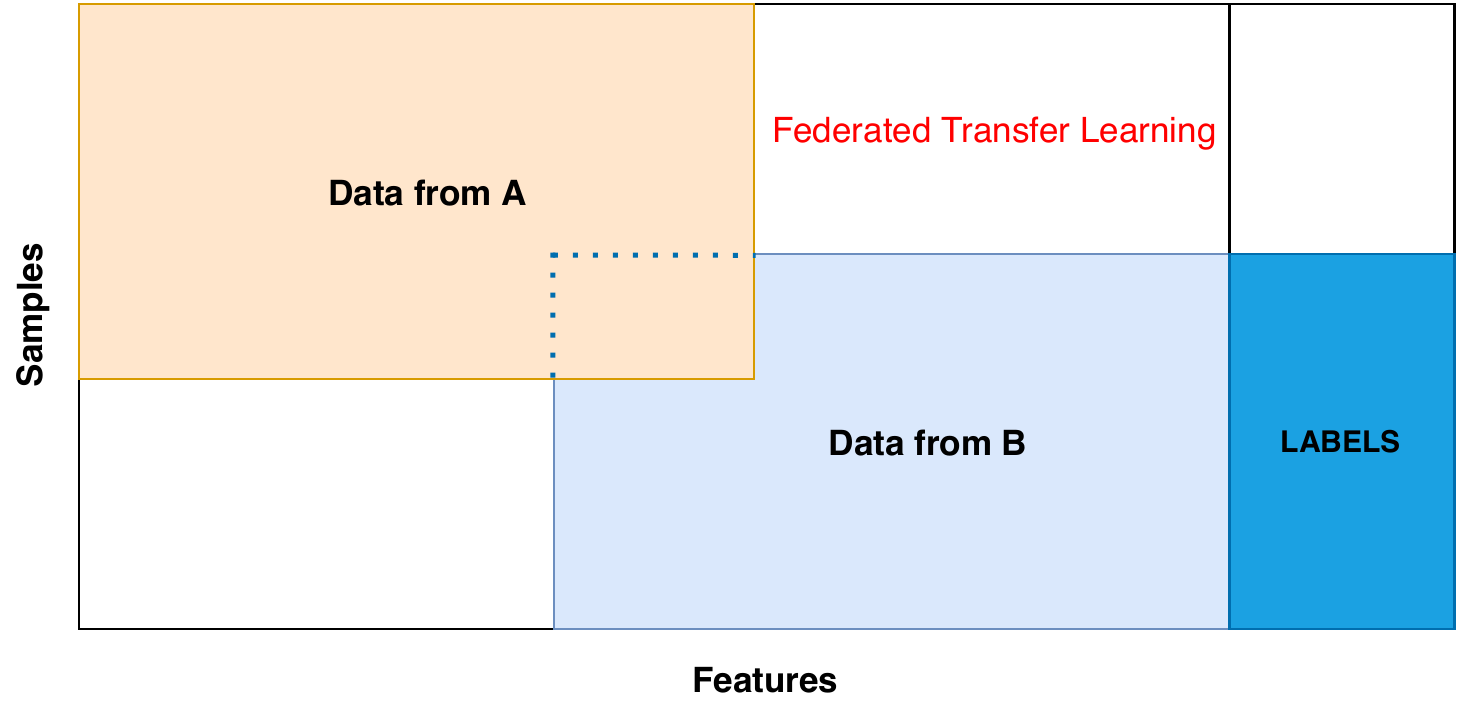}
\caption{Federated Transfer Learning \cite{yang2019federated}}
\label{figureFTL}
\end{figure}

Federated transfer learning is a special case of federated learning and different from both horizontal and vertical federated learning. In federated transfer learning, two datasets differ in the feature space. This applies to datasets collected from enterprises of different but similar nature. Due to the differences in the nature of business, such enterprises share only a small overlap in feature space. This is also applicable to the enterprises set up far in globe. Thus in such scenarios, datasets differ both in samples and in feature space.
Transfer learning techniques aim to build effective model for the target domain while
leveraging knowledge from the other (source) domains. A typical architecture of federated transfer learning is shown in
Figure \ref{figureFTL}. Considering two parties A and B, where there is only a small overlap in feature space and sample space between A and B, a model learned on B is transferred to A by leveraging small overlapping data and features. We recall that horizontal federated learning is used when there is large overlap in the feature space between datasets and vertical federated learning is used when there is large overlap in user/sample space between datasets. In contrast to them, FTL is used when there is small overlap in both feature space and sample space (shown by dotted box in Figure \ref{figureFTL}). FTL ingests a model trained on source domain samples and feature space. Subsequently FTL orients the model for reuse in target space such that model is used for non-overlapping samples leveraging the knowledge acquired from source domain non-overlapping features. Thus FTL covers the region in right upper corner of the Figure \ref{figureFTL} by transferring knowledge from non-overlapping features from source domain to the new samples in the target domain. The ability to use the transferred on non-overlapping data in A
makes FTL different from vertical transfer learning.

\section{Use Cases of FTL}
\label{sectionFTLCaseStudies}

\subsection{Wearable Healthcare}
\label{sectionWearableHealthcare}
\begin{figure}[ht]
\centering
\includegraphics[width=0.8\linewidth,keepaspectratio]{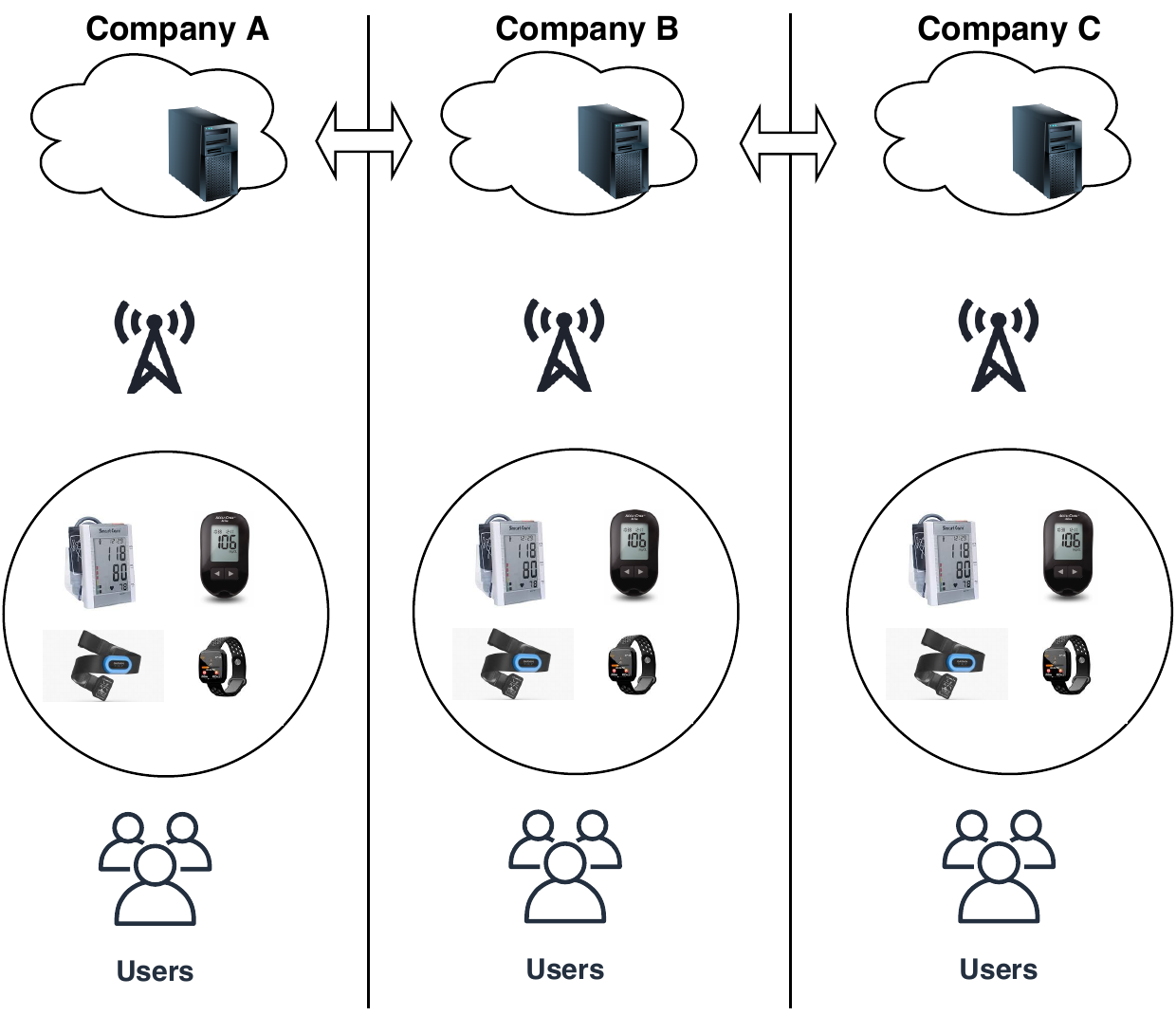}
\caption{Architecture of the wearable healthcare system}
\label{fig:wearable}
\end{figure}

Wearable devices are fast becoming part of everyday life for patients and healthcare providers. 
Multiple features and functionalities of wearable devices include remote patient monitoring, tracking and collecting data, enhancing everyday health and lifestyle patterns, detecting chronic conditions, among others. 
Healthcare data, however, are usually fragmented and private making it difficult to generate robust results across populations. A typical architecture of wearable healthcare system is shown in Figure~\ref{fig:wearable}. It can be seen that the users use their wearable healthcare devices to measure various health related parameters they are concerned about. The user's data generated by healthcare devices often exist in the form of isolated islands. For further assessment and analysis of results obtained the collected data is uploaded to the remote cloud based server~\cite{sun2020gait}. 

The architecture presents several critical challenges that hinders the development of effective analytical approaches for the generalizing of healthcare data. The adequate posture of healthcare data has several root causes including, (i) regulatory restrictions---lack of acquisition of massive user data, (ii) security and privacy---restricts sharing of data that exist in the form of isolated islands, and (iii) personalizing issue---the process of training machine learning model lacks personalization~\cite{chen2020fedhealth}.

\subsection{EEG signal classification}
In addition to Section \ref{sectionWearableHealthcare}, another example of usage of federated transfer learning in healthcare domain is the electroencephalographic (EEG) signal classification \cite{ju2020federated}.
Brain-Computer Interface (BCI) systems aim to decode participants' brain states. The success of deep learning based BCI models for classification of EEG recordings is restricted by lack of large EEG datasets. Due to the 
privacy concern and high data collection expenses, EEG-BCI
data is present in the form of multiple small datasets owned by different entities across the globe.
\par
Towards this, Ju \textit{et. al.} \cite{ju2020federated} proposes a method where the EEG data is represented as the spatial covariance matrix and is subsequently fed to a deep learning based federated transfer learning architecture. It is assumed that the architectures of each user's deep classifier is same. 
Federated averaging method \cite{mcmahan2016federated} is adopted to aggregate models from different users. A server-client setting is used where a server acts as the model aggregator. In each round, the updated local models are sent to the server and server sends back the updated global model after aggregation. When a client receives the global model, it updates the model with its local data. This work \cite{ju2020federated} clearly demonstrates that use of domain adaptation in 
federated learning architecture boosts EEG classifier performance.

\subsection{Autonomous Driving}
\begin{figure}[ht]
\centering
\includegraphics[width=0.8\linewidth,keepaspectratio]{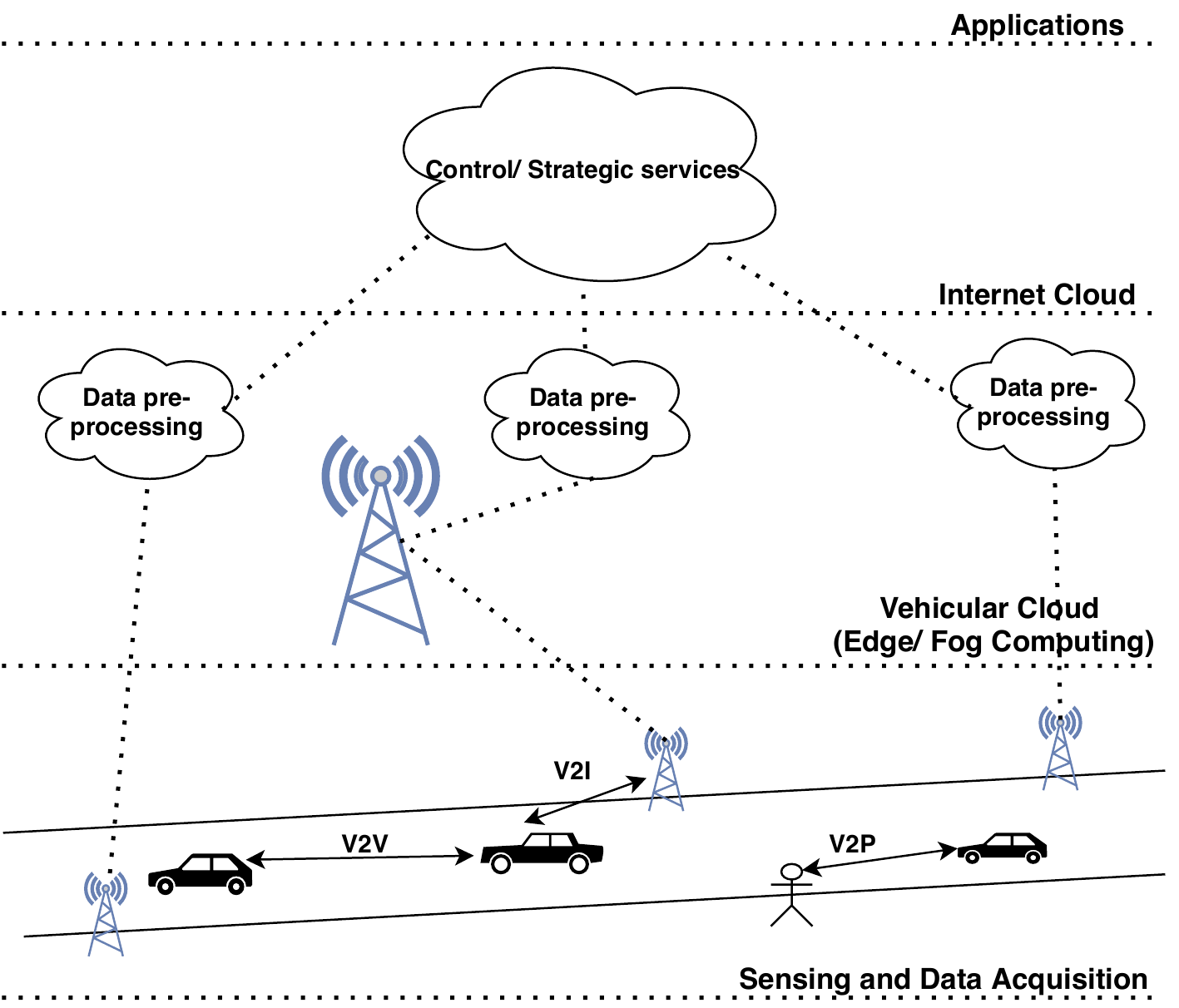}
\caption{Layered architecture of autonomous driving}
\label{fig:driving}
\end{figure}

Autonomous driving normally refers to self-driving vehicles or transport systems that move without the intervention of a human driver. As seen in Figure~\ref{fig:driving} autonomous driving technology is a complex integration of technologies including sensing, perception, and decision. The cloud platform (vehicular and Internet) provides data storage, simulation, high definition (HD) map generation, and deep learning model training functionalities. Autonomous vehicles are mobile systems, and autonomous driving clouds provide some basic infrastructure supports including distributed computing, distributed storage, and heterogeneous computing ~\cite{liu2017implementing}. On top of this infrastructure, essential services can be implemented in the form of applications to support autonomous vehicles.

The dynamic nature of the autonomous driving environment and the uncertainty of real-life scenarios makes autonomous driving as a special use-case for FTL ~\cite{liang2019federated}.

\subsection{Image steganalysis}
Image steganography is the technique of hiding information in the digital image without compromising its visual aesthetics. Image steganalysis is a counter technique to image steganography. It aims to detect the hidden information in
the digital images. Towards this, it extracts and analyzes the steganographic
features generated by image steganographic algorithms. There is a lack of data for training steganalysis methods due to the unwillingness of sharing data among the steganographers. Furthermore,  different data owners may
have different preference for steganographic algorithms and cover images. The traditional image
steganalysis algorithms cannot account for this personalized preference. 

\begin{figure}[ht]
\centering
\includegraphics[width=0.8\linewidth,keepaspectratio]{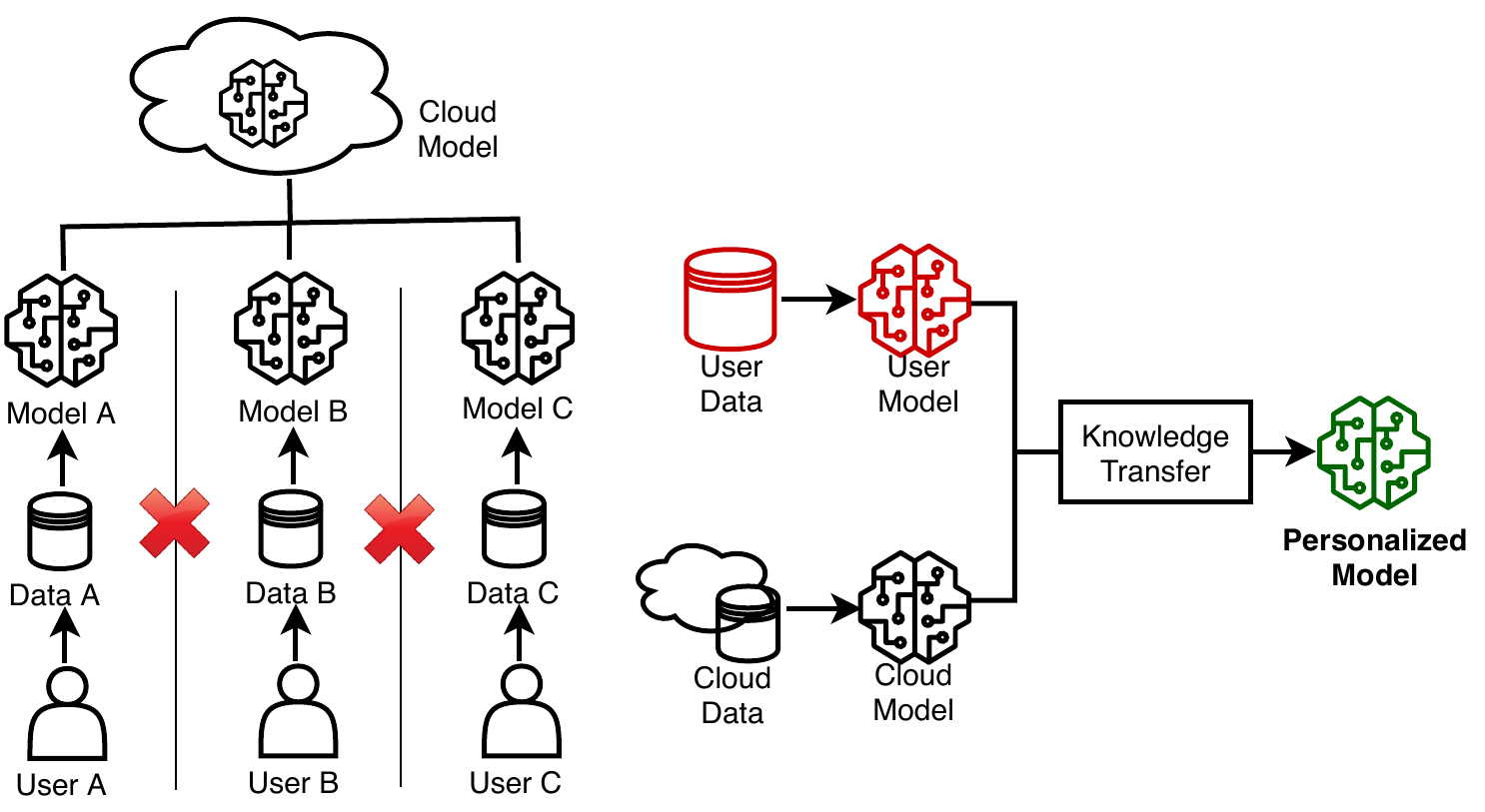}
\caption{Users A, B, and C denote steganographic images owner and they do not share any information (left). For personalized model training, each user individually interacts with the cloud model (right).}
\label{figSteg}
\end{figure}

To overcome these challenges, Yang \textit{et. al.} \cite{yang2020fedsteg} proposed federated transfer learning framework for image steganalysis (FedSteg), as shown in
Figure \ref{figSteg}. In the proposed framework, different users have their own data (stego and cover images). The users do not leak the data to each other. Instead it is assumed that there is a cloud model and personalized local model for each user. 
The
cloud model is trained with the data (cover and stego images) on the
cloud-side. Then the cloud model is distributed to the users. Each user
trains their local model with local data. The trained user models are sent back to the cloud side to help it train a new cloud model. This step only shares encrypted model parameters. In this fashion, each user can keep performing personalized training by consolidating the new cloud model with their previous model and its data \cite{yang2020fedsteg}. However, there will be distribution discrepancy between the
cloud and user data. To mitigate the distribution discrepancy, transfer learning is performed to make the model more suitable for user. Yang \textit{et. al.}
\cite{yang2020fedsteg} uses a CNN based deep transfer learning to train personalized model for each user.

\section{Privacy in FTL}
\label{sectionFTLPrivacy}
Machine learning relies on the availability of vast amounts of data for training. However, in real life scenarios (as seen in Section~\ref{sectionFTLCaseStudies}), data are mostly scattered
across different organizations and cannot be easily integrated due to many legal and practical constraints. Federated transfer learning (FTL) helps to improve statistical modeling under a data federation. The data federation as in case of FTL allows knowledge to be shared without compromising user privacy, and enables complimentary knowledge to be transferred in the network. Thus allowing the target domain party to develop a more flexible and powerful model by leveraging rich labels from source domain party ~\cite{liu2020secure}.

The broad application of FTL is currently hindered by limited dataset availability for algorithm training and validation, due to the absence of technical and legal approaches to protect user's privacy. FTL has strict privacy preservation requirements, therefore, to prevent user privacy compromise while promoting scientific research on large datasets, the implementation of technical solutions and development/implementation of legal frameworks to simultaneously address the demands for data protection and utilization is mandatory. 

Privacy-preserving FTL typically involves multiple
parties with emphasis on security guarantees to perform machine learning. Here, we present an overview of the necessary privacy-preserving approaches and techniques in context of FTL~\cite{kaissis2020secure, gao2019privacy}.

\begin{itemize}
    \item \textbf{Privacy by Design} FTL designed from the ground up with privacy in mind. The idea is taking into account privacy, throughout the FTL development process. The principles of privacy by design may be applied to all types of sensitive data and the strength of the implemented privacy measure must be dependent on the sensitivity of subject data. For example, processing of only necessary data, storage of data for minimal period, and limited accessibility. Optimal privacy preservation requires implementations that are secure by default and require minimal or no data transfer and provide theoretical and/or technical guarantees of privacy~\cite{cavoukian2011privacy,kaissis2020secure}.
    
    \item \textbf{Anonymization and pseudonymization} The former refers to the removal of personally identifiable information from a dataset (e.g., removing information related to age and gender), whereas, the later refers to replacement of personally identifiable information in a dataset with a synthetic entry with separate storage of the linkage record. For example, in case of health insurance companies wishing to reduce financial risk, re-identification of patient records are a lucrative target. Recently, data mining companies are adopting large-scale re-identification attacks and the sale of re-identified medical records as a business model~\cite{tanner2017our}. Keeping that into account, the use of naive anonymization or pseudonymization alone must therefore be viewed as a technically insufficient measure against identity inference~\cite{kaissis2020secure}.
    
    \item \textbf{Differential privacy} The alteration of a dataset to obfuscate individual data points while retaining the ability of interaction with a data within a certain scope (privacy budget) and of statistical analysis. The approach can also be applied to algorithms. For example, randomization of data to omit relationships between individuals and respective data entries. It provides privacy preservation against membership-inference attack in the model inference stage ~\cite{dwork2006calibrating}. However, during model training FTL approaches based on differential privacy are vulnerable to privacy leakage among the participants  ~\cite{wang2018differentially, yao2018differential}.
    
    \item \textbf{Homomorphic encryption} A cryptographic technique that preserves the ability to perform mathematical operations on data as if it was unencrypted i.e., plain text. For example, performing neural network computations on encrypted data without the need of first decrypting it. Homomorphic encryption is studied for private federated logistic regression on vertically partitioned data ~\cite{hardy2017private}. More recently,~\cite{nikolaenko2013privacy} proposed a privacy-preserving linear regression on horizontally partitioned data using homomorphic encryption and Yao’s Garbled Circuits~\cite{yakoubov2019gentle}.
    
    \item \textbf{Secure multi-party computation} The technique is based on splitting data among collaborating entities to perform joint computation but prevents any collaborating entity from gaining knowledge of the data. For example, identifying the common patients among two hospitals without disclosing the respective hospital patient's list. Earlier works ~\cite{kantarcioglu2004privacy, wan2007privacy, yao1982protocols} mostly focus on approaches based on multi-party computation. SecureML ~\cite{mohassel2017secureml}, a privacy-preserving protocol combining secret-sharing ~\cite{shamir1979share} and Yao’s Garbled Circuit~\cite{yakoubov2019gentle}, is considered as the state-of-the-art protocol for linear regression, logistic regression, and neural networks. SecureNN~\cite{wagh2018securenn} is also proposed using a multi-party protocol for efficient neural network training.
    
    \item \textbf{Hardware security implementations} The approach to assure data and algorithm privacy by utilizing specialized hardware. For example, in the form of secure processors or enclaves implemented in mobile device~\cite{apple}. Due to the rising significance of hardware-level deep learning implementations, e.g., tensor processing units~\cite{google}). It is likely that such system-based privacy guarantees built into edge hardware will become more common, e.g., trusted execution environments~\cite{kaissis2020secure}.
    
\end{itemize}

\section{Machine learning in FTL} 
\label{sectionFTLMachineLearning}
Most works on FTL \cite{chen2020fedhealth, ju2020federated, yang2020fedsteg} have adopted variants of deep learning as the architecture for FTL. 
\par
In \cite{chen2020fedhealth}, the deep model is formed using a series of 1D convolution layers and fully connected layers. The model also consists of other auxiliary layers like pooling. Softmax layer is used for classification. A simplified schema of the deep model is shown in Figure~\ref{figureDeepModelWearableHeath}. During the model from user to server, the convolution layers are kept frozen and the fully connected layers are updated to learn user and task-specific parameters. 
\par
\tikzset{
    between/.style args={#1 and #2}{
         at = ($(#1)!0.5!(#2)$)
    }
}
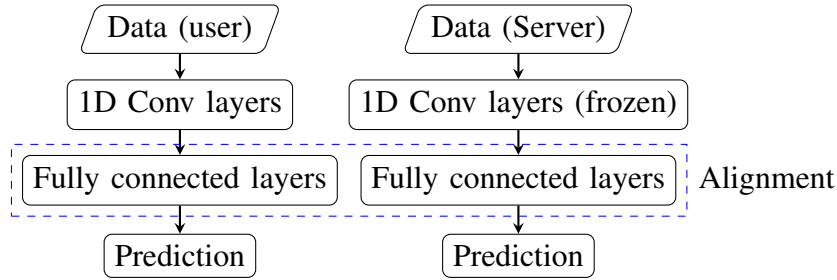
\begin{figure*}
 \centering
 \begin{tikzpicture}
\node (1a) [rounded corners=3pt,draw, io, align=center] {Data (user)};
\node (1b) [rounded corners=3pt,draw, io, align=center, right of=1a,xshift=3.5cm] {Data (Server)};
\node (2a) [rounded corners=3pt,draw, align=center, below of=1a, yshift=-0.005cm] {1D Conv layers};
\node (2b) [rounded corners=3pt,draw, align=center, below of=1b, yshift=-0.005cm] {1D Conv layers (frozen)};
\node (3a) [rounded corners=3pt,draw, align=center, below of=2a, yshift=-0.005cm] {Fully connected layers};
\node (3b) [rounded corners=3pt,draw, align=center, below of=2b, yshift=-0.005cm] {Fully connected layers};
\node (4a) [rounded corners=3pt,draw, align=center, below of=3a, yshift=-0.005cm] {Prediction};
\node (4b) [rounded corners=3pt,draw, align=center, below of=3b, yshift=-0.005cm] {Prediction};

\draw [arrow] (1a) -- (2a);
\draw [arrow] (1b) -- (2b);
\draw [arrow] (2a) -- (3a);
\draw [arrow] (2b) -- (3b);
\draw [arrow] (3a) -- (4a);
\draw [arrow] (3b) -- (4b);

\node (ancillary1) [right of=3b,xshift=2.3cm] {Alignment};
\node[draw, blue, dashed,fit= (3a) (3b)]{};

\end{tikzpicture}
\caption{The deep learning model for FTL \cite{chen2020fedhealth}}
\label{figureDeepModelWearableHeath}
\end{figure*}

A similar framework as above is used in FedSteg \cite{yang2020fedsteg}. The model consists of 9 convolution layers followed by fully connected layer. While transferring model from user to cloud/server, the convolution layers are kept frozen, while the fully connected layer is used for model transfer. For feature alignment between source and target, correlation alignment (CORAL) loss is used, which adapts the second order feature statistics \cite{yang2020fedsteg}. The work in \cite{ju2020federated} uses both classification loss and domain loss that is implemented using maximum-mean discrepancy \cite{gretton2007kernel}.
\par
In federated autonomous driving \cite{liang2019federated}, reinforcement learning is used. Reinforcement learning \cite{li2017deep}, different from other machine learning techniques, can be considered as a collection of observation and action spaces. In \cite{liang2019federated}, each RL agent (user) is trained individually. Following this, knowledge of the distributed RL agents is distributed. Following that an online transfer process is performed to make alignments on the observations and actions. 
\par
In FTL, training with heterogeneous data may present additional challenge, e.g.,
not all client data distributions may be adequately captured by the model. Furthermore, quality of a particular local data partition may be significantly different from the rest. To overcome these challenges, Dimitriadis \textit{et. al.} \cite{dimitriadis2020federated} presented a dynamic gradient aggregation (DGA) method which weights the local gradients during aggregation step. 
\par
Inspite of success of the methods discussed above, most of them do not provide an in-depth analysis of how different feature spaces can be handled in different users. Moreover, most of the above works are restricted to limited number of users. They are based on simplistic assumptions that convolution layers learn shared feature while fully connected layers learn domain-specific/ task-specific features. While correct, such assumptions do not fully exploit the tremendous progress made by deep domain adaptation community
\cite{wilson2020survey}. Furthermore, there are very few works that investigates the complicated machine learning issues that can arise in the heterogeneous FTL setting \cite{dimitriadis2020federated}.
\section{FTL datasets} 
\label{sectionFTLDataset}
Most works on FTL have adopted existing machine learning datasets and modified them as per the requirement of FTL.
\par 
Fedhealth \cite{chen2020fedhealth} used human activity recognition dataset - UCI Smartphone \cite{anguita2012human}, consisting of 6 activities collected from 30 users within an age bracket of 19-48 years. 
To adapt the dataset for FedHealth,
5 subjects were regarded as isolated
users (who cannot share data) and the remaining were used to train cloud model.
\par
Ju \textit{et. al.} \cite{ju2020federated} used PhysioNet
EEG Motor Imagery (MI) Dataset \cite{schalk2004bci2000}. The MI dataset is recorded from 109 subjects. Their experiments \cite{ju2020federated} use 5-fold cross-validation settings with 4 folds being
used for training. Fedsteg \cite{yang2020fedsteg} used two different image datasets. Gao \textit{et. al.} \cite{gao2019privacy} conducted experiments from several 
public datasets from UCI repository \cite{asuncion2007uci}.
\par
Differently from \cite{chen2020fedhealth}, \cite{gao2019privacy} and \cite{ju2020federated},
Liang \textit{et. al.} \cite{liang2019federated} conducted real-life experiments on RC cars and Airsim.  

\section{Conclusions} 
\label{sectionConclusion}
Data isolation and data privacy concerns pose significant challenge for applying artificial intelligence techniques in real life applications. 
Moreover, features and users generally vary from organization to
organization.
In the last few years, federated learning, especially FTL has emerged as
a potential solution to overcome the above challenges. In this work, we analyzed the concept of FTL and its applications in several practical domains. Furthermore, we did a detailed review of the machine learning techniques used in FTL. Our analysis shows that while FTL has explored different machine learning techniques, there is still potential of using more sophisticated machine learning techniques along with FTL. Adversarial techniques may provide a  promising direction for further strengthen FTL. While FTL has been used in some practical applications, number of such applications are still few. Moreover, most of the datasets used in the FTL works are basic machine learning datasets and not tailored to real life scenarios.  FTL holds the promise to break the barrier between different enterprises. However, there is scope of further advancement, by incorporating advanced machine learning techniques in FTL and applying FTL to more practical applications.

\ifCLASSOPTIONcaptionsoff
  \newpage
\fi

\bibliographystyle{IEEEtran}
\bibliography{federatedTransferLearning}









\end{document}